\documentclass[runningheads]{llncs}

\usepackage{eccv}

\usepackage{eccvabbrv}

\usepackage{graphicx}
\usepackage{booktabs}

\usepackage[accsupp]{axessibility}  

\usepackage{hyperref}

\usepackage{orcidlink}
\usepackage{booktabs}
\usepackage{multirow}
\usepackage{multicol}
\usepackage{titletoc}
\usepackage{amsmath}  
\usepackage{xcolor}         
\usepackage{colortbl}
\usepackage{wrapfig}
\begin{document}

\title{Diffusion Models are Open-World Affordance Learners: Leveraging Generative Priors for 3D Affordance Learning} 
\titlerunning{Diffusion Models are Open-World Affordance Learners}

\author{
Hanqing Wang\inst{1,}$^*$ \orcidlink{0009-0007-3329-2731} \and
Zhenhao Zhang\inst{2, }$^*$ \and
Kaiyang Ji\inst{2, }$^*$ \and
Mingyu Liu\inst{3} \and
Wenti Yin\inst{5} \and
Yuchao Chen\inst{5} \and
Zhirui Liu\inst{2} \and
Xiangyu Zeng\inst{4} \and
Tianxiang Gui\inst{2} \and
Hangxing Zhang\inst{2} \and
Jiahao Yuan \inst{1}\and
Zhiqing Cui\inst{1} \and
Jiaxin Liu\inst{2} \and
Zhiyuan Ma\inst{5,\dagger} \and
Hui Xiong\inst{1,6,\dagger}
}

\authorrunning{Hanqing Wang et al.}

\institute{
The Hong Kong University of Science and Technology, Guangzhou, China\and
ShanghaiTech University, Shanghai, China\and
Zhejiang University, Hangzhou, China\and
Nanjing University, Nanjing, China\and
Huazhong University of Science and Technology, Wuhan, China\and
The Hong Kong University of Science and Technology, Hong Kong, China
\email{hwang201@connect.hkust-gz.edu.cn}
\\
\noindent $^*$ Equal contribution.
\noindent $\dagger$ Corresponding author.
}

\maketitle
\begin{center}
    \captionsetup{type=figure}
    \includegraphics[width=1\linewidth]{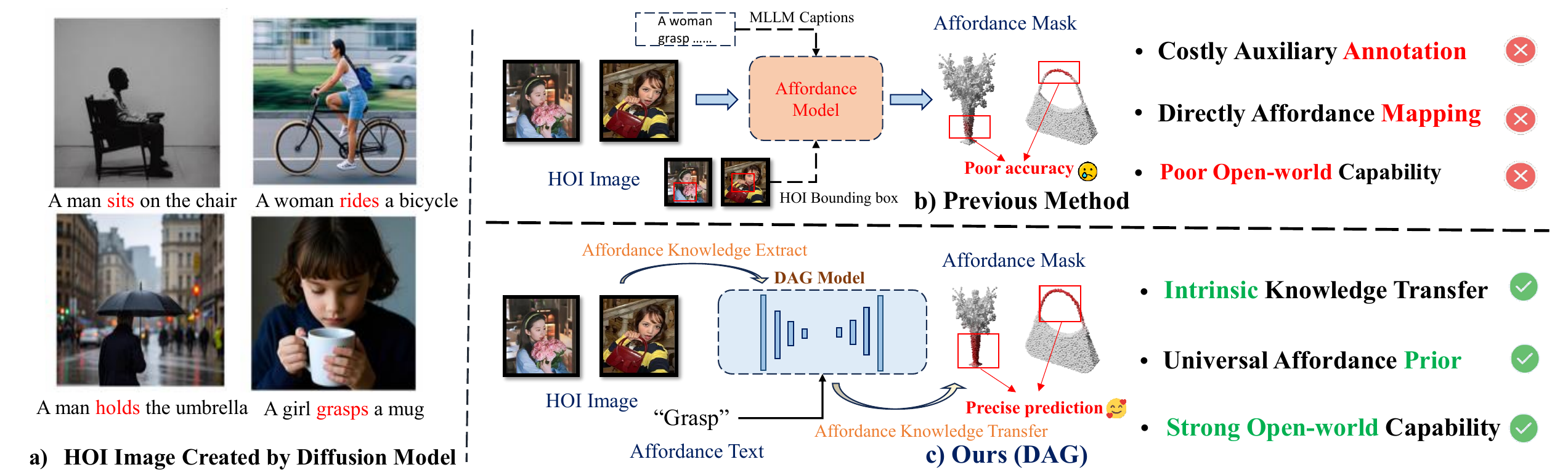}
    \captionof{figure}{Motivation: Text-to-Image Diffusion model can understand how people interact with objects. It has an awareness of affordance and can generate reasonable Human-Object Interaction (HOI) images \textit{(Left)}. Motivated by this, we would like to find a way to transfer this rich affordance knowledge into 3D affordance grounding \textit{(Right)}.}
\end{center}
\label{figvis}

\begin{abstract}

3D affordance grounding aims to understand how diverse objects can be manipulated, making it a cornerstone of embodied interaction. However, prior works struggle to generalize to out-of-distribution, open-world scenarios, leaving a critical gap between limited dataset performance and real-world application needs. Inspired by the saying: \textit{\textbf{``What I can not create, I do not understand''}}, we find generative models can generate semantically valid HOI images, which indicates inherent encoding of affordance concepts. Building on this insight, we propose DAG, the first innovative diffusion-based 3D affordance grounding framework that extracts general affordance knowledge from text-to-image diffusion models for 3D affordance prediction. Specifically, we extract the affordance priors from a diffusion model to encode HOI priors, and design an affordance block with a multi-source affordance decoder for dense 3D affordance prediction. Extensive experiments show that DAG consistently outperforms state-of-the-art methods and exhibits strong open-world generalization, even in the challenging one-shot setting. The code of our method is released on \textcolor{blue}{\textit{https://github.com/hq-King/DAG}}.

\end{abstract}

\section{Introduction}
\label{sec:intro}



3D affordance grounding is essential for embodied intelligence, as it unifies semantic perception with actionable interaction and serves as a fundamental bridge between perception and manipulation. While conventional object recognition focuses on identifying \emph{what} an object is, affordance grounding addresses the more fundamental problem of understanding \emph{how} an object can be functionally used. This capability is indispensable for intelligent agents that aim not only to perceive the environment but also to interact with it purposefully. For instance, given a natural-language query such as “Which region of the cup enables stable grasping?” or a visual demonstration of human-object interaction (HOI), a reliable affordance-grounding model should precisely identify the handle as the functional interaction region.


Existing approaches~\cite{gao2025learning,yang2023grounding} investigate learning affordance knowledge from HOI images and language instructions via diverse training paradigms.
However, they frequently require extra supervision (e.g., bounding boxes or affordance-centric captions) and still struggle to capture consistent, generalizable regularities of real-world interactions.
As a result, they only partially capture affordance semantics and do not leverage broad, transferable world knowledge, both of which are crucial for robust open-world 3D affordance grounding.


Recently, diffusion models trained on internet-scale data have revolutionized image synthesis~\cite{balaji2022ediff,rombach2022high,ramesh2022hierarchical}. By learning the distribution of real-world visual data, these models acquire strong world knowledge and can generate highly realistic and physically plausible HOI images~\cite{yang2023boosting}. Inspired by Feynman’s insight—\textbf{\textit{“What I cannot create, I do not understand”}}, we find that generative models, especially diffusion models, inherently contain affordance knowledge within their generative priors. Generation and understanding are closely coupled: a model that can reliably synthesize coherent HOI images must implicitly capture the affordances underlying human-object interactions, such as admissible manipulation modes and the spatial constraints that make them feasible.

\begin{figure}[t]
    \centering
    \includegraphics[width=\linewidth]{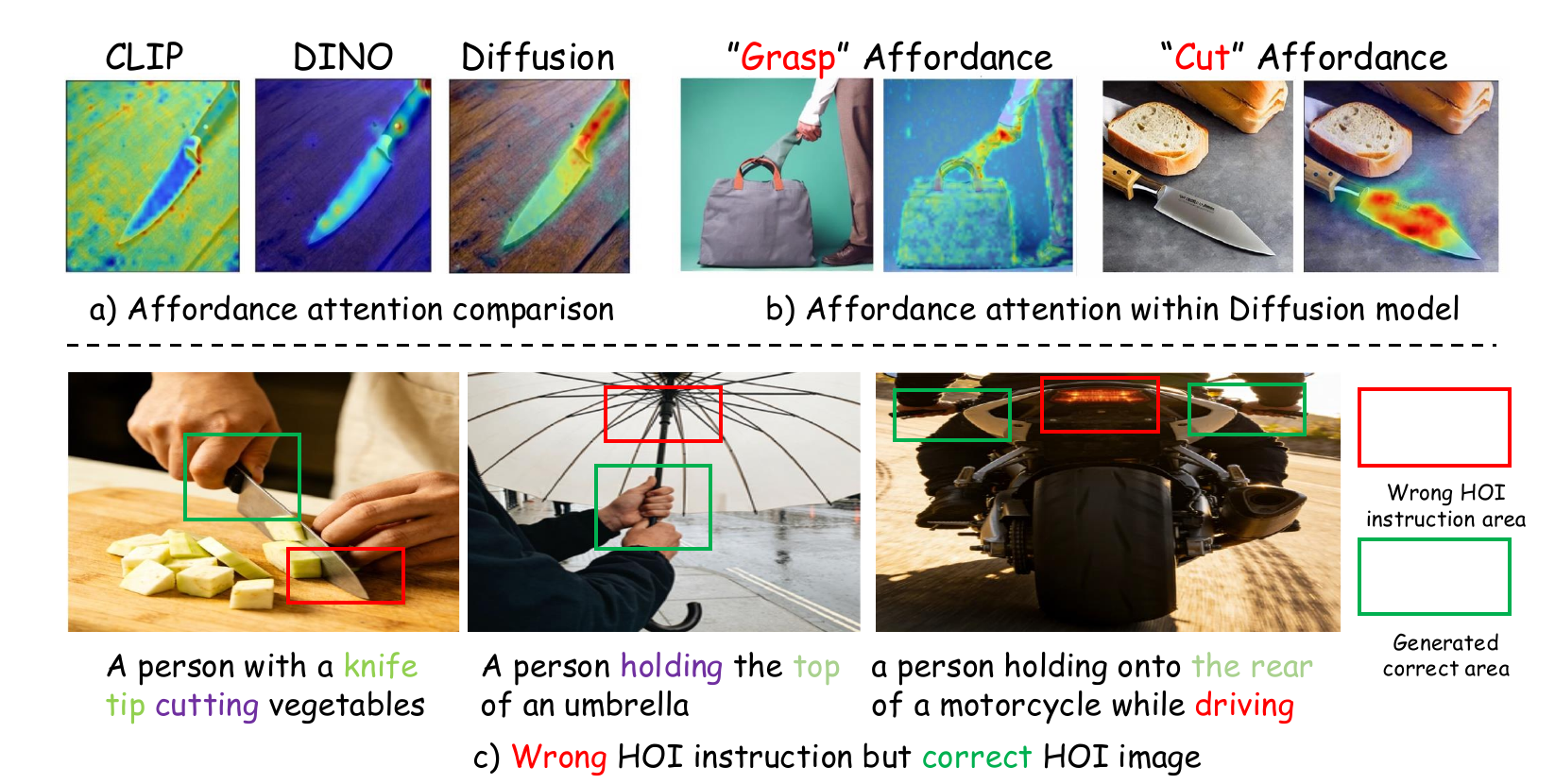}
    \caption{We conduct some toy experiments here. As shown in a), compared with CLIP and DINO, the diffusion model can understand the affordance concept of "hold" better and locate the accurate affordance region. b) shows that the inherent affordance knowledge in the diffusion model can reveal the affordance region on both the ego and exo image. c) When the interaction area indicated by the instruction we give is incorrect, the diffusion model can still generate correct and reasonable interaction images, which indicates that it truly understands \textit{``how to interact''} and \textit{``where to interact''} represented by affordance internally, rather than simply mapping data distribution.}
    \label{affordance_attention}
\end{figure}

Motivated by this, we wonder whether a large-scale text-to-image diffusion model can serve as a universal affordance learner for open-world affordance concepts, and whether its implicit knowledge can be effectively transferred to 3D affordance grounding. As visualized by the attention maps in Figure~\ref{affordance_attention}, diffusion models already demonstrate a strong ability to implicitly capture affordance cues embedded in HOI imagery. This motivates us to develop a novel framework that extracts and transfers rich generative priors from diffusion models to improve 3D affordance grounding, enabling more generalizable, robust, and physically consistent functional understanding.

We propose \textbf{DAG}, the first diffusion-based 3D affordance grounding model that leverages the rich affordance knowledge encoded in diffusion models for 3D affordance grounding. An overview of our approach is illustrated in Fig. \ref{pipeline}. At a high level, we use a pre-trained, frozen text-to-image diffusion model as a feature extractor: given an HOI image and its affordance text, we obtain multi-level internal diffusion features. Building on the extracted diffusion features, the affordance block integrates textual and visual cues to form unified affordance features. Finally, a multi-modal affordance decoder combines diffusion-derived affordance knowledge with point features extracted by a frozen point encoder to predict dense 3D affordance masks.

To summarize, our contributions are as follows:
\begin{itemize}
    
    \item We verify that diffusion models inherently encode rich affordance knowledge. Based on this observation, we propose a novel framework named DAG, which unlocks the implicit affordance knowledge within diffusion models and transfer it to 3D affordance grounding.

    \item To better leverage affordance knowledge extracted from text-to-image diffusion models for 3D affordance grounding, we introduce an affordance block and a multi-modal affordance decoder, which effectively enhance the model’s 3D grounding capability.
    
    \item We implement extensive experiments to demonstrate the effectiveness of our learning pipeline and observe noticeable gains over baselines with strong generalization capability, which highlights the effectiveness and adaptability of our approach in real-world applications.
\end{itemize}
\section{Related Work}

\subsection{Affordance Grounding}

The concept of affordance, popularized by psychologist James Gibson~\cite{gibson1977theory}, refers to the actionable possibilities that an environment or object offers to an agent. A substantial body of work has sought to extract and model affordance knowledge from diverse data sources. In robotics, affordance learning enables robots to interact effectively with complex and dynamic real-world environments~\cite{nguyen2023open}. In particular, several approaches leverage affordances to relate objects, tasks, and manipulations, supporting robust robotic grasping~\cite{tang2025affordgrasp,ji2025robobrain,li2024manipllm,chu20253d,liu2026affordance2action}. Other lines of work learn affordances from data that can be deployed on real robots, such as human-teleoperated interaction trajectories.
In computer vision, early studies primarily focused on 2D affordance detection with convolutional neural networks~\cite{xuweakly,qian2024affordancellm,li2024one,wang2025affordance}, aiming to detect or segment action-related regions from 2D visual inputs. However, 2D affordance representations do not readily extend to the 3D spatial configurations required for physical interaction in the real world. To address this limitation, 3D AffordanceNet~\cite{deng20213d} introduced a benchmark for affordance grounding on 3D point clouds. Building on this benchmark, OpanAD~\cite{nguyen2023open} successfully exploits the semantic relationships between affordances by simultaneously learning the affordance text and the point feature. IAGNet~\cite{yang2023grounding} and MIFAG~\cite{gao2025learning} pursue 3D affordance grounding from reference images. GREAT introduce affordance-centric captions to help with the affordance grounding. More recently, researchers have explored extending the reasoning capabilities of MLLMs~\cite{yu2025seqafford,wang2026videoafford,zhang2025openhoi} to affordance grounding. LMAffordance~\cite{zhu2025grounding} applies a vision-language model to fuse 2D and 3D spatial features with semantic features. Seqafford~\cite{yu2025seqafford} further explores how to transfer the sequential reasoning capability of large models to affordance grounding. Nevertheless, these methods typically map reference images or language instructions directly to 3D structures. As a result, they struggle to capture shared affordance regularities across diverse human–object interaction images and to model affordance knowledge in a transferable manner. In contrast, we propose leveraging diffusion models to learn more universal affordance knowledge: our observations suggest that diffusion models act as open-world affordance learners, implicitly encoding strong affordance concepts and broad world knowledge. In this paper, we investigate how to transfer and elevate this capability for 3D affordance grounding.

\subsection{Image-Point Cloud Cross-Modal Learning}
Multimodal learning\cite{wang2026sdevalsafetydynamicevaluation,zhang2025hoidr1reinforcementlearningopenworld}, which aims to integrate and exploit complementary information from heterogeneous data sources, plays a critical role in enhancing the perception and understanding capabilities of intelligent systems. The individual limitations of each sensor type, such as LiDAR's lack of color information and cameras' inability to directly measure depth, highlight the necessity of integrating these data sources for a more complete scene understanding\cite{yao2024cmr}. Image-point cloud cross-modal learning has emerged as a promising research direction to overcome the limitations of relying on either modality alone~\cite{liu2025cross}.To address the challenges of modality discrepancy and effective information fusion, many works have explored joint representation learning and fusion strategies~\cite{afham2022crosspoint,aiello2022cross,chen2023pointdc,choo2024supervised,dong2023autoencoders,qi2023contrast,yan2022let,zhou2024pointcmc,zhang2026unihmunifieddexteroushand,liu2025dream3davatartextcontrolled3davatar}. In parallel, some studies focus on explicitly modeling the cross-modal correspondences between visual and geometric features to guide the interaction process~\cite{mao2025dmf,xu2024explicitly,yuan2023pointmbf}. In addition, recent research has also explored the use of external priors or generative models to compensate for incomplete geometry~\cite{du2024cdpnet,kasten2023point,ma2022unsupervised,li2025genpc,zhou2025fine}. Recently, several works in affordance learning have leveraged image-point cloud cross-modal learning to achieve 3D object affordance grounding \cite{gao2025learning,yang2023grounding}. These approaches align interaction cues from 2D images with 3D geometric structures, enabling the localization in 3D space. By employing cross-modal alignment and contextual modeling, such methods enhance the model’s ability to perceive and reason about object affordances across modalities. However, these methods utilize CNN-based ways to directly map the connection between images and 3D structures, while our method aims at unlocking the internal affordance knowledge within the text-to-image diffusion model and transferring this into affordance grounding. Benefiting from its powerful generative prior and holistic scene understanding, the diffusion-based features offer richer semantic cues and more precise alignment with 3D geometric structures, leading to enhanced cross-modal representation and affordance reasoning.

\subsection{Diffusion Models for Visual Tasks} 
Diffusion probabilistic models (DPMs)~\cite{dhariwal2021diffusion,tang2023emergent} have rapidly become a leading paradigm for high-fidelity image synthesis by reversing a fixed Markovian noising process. Recent works~\cite{xu2023open,dhariwal2021diffusion,ye2023affordance,kim2024beyond,kim2025david,zhang2025h2oflow,ma2023lmdfasterimagereconstruction,ma2024efficientdiffusionmodelscomprehensive,ma2024neuralresidualdiffusionmodels} show that the U-Net backbones of diffusion models encode rich internal representations that can be repurposed for a wide range of downstream tasks. Learned through the denoising process, these intermediate features capture both global and local semantics, and have proven effective for image segmentation~\cite{baranchuk2021label}, depth estimation~\cite{ke2024repurposing}, and object recognition~\cite{chen2023diffusiondet}. For example, \cite{li2023your} demonstrates that diffusion latent spaces can support image classification and related tasks without fine-tuning. Similarly, \cite{tian2024diffuse} exploits diffusion activations for semantic segmentation, highlighting their potential as reusable and flexible representations. \cite{saharia2022photorealistic} further apply diffusion features to cross-modal learning, such as text-to-image alignment, while \cite{samuel2024s} show improvements in self-supervised objectives (e.g., image retrieval), extending diffusion representations beyond generation.
In contrast to prior efforts that primarily leverage diffusion internals for 2D vision, we are the first to systematically extract and transfer affordance knowledge from frozen diffusion representations into a 3D affordance grounding framework, enabling accurate prediction of interaction regions on point clouds even with minimal 3D supervision.\textbf{}

\section{Method}

\subsection{Overview}
Following previous works, our goal is to anticipate the affordance regions on the point cloud corresponding to the reference human-object interaction (HOI) image and the affordance text. Given a sample $(P, I, T, y)$, where $P$ is a point cloud with coordinates $P \in \mathbb{R}^{N \times 3}$, $I \in \mathbb{R}^{3 \times H \times W}$ is an RGB image, $T$ represents the affordance text, and $y$ is the affordance category label, DAG utilizes the rich affordance knowledge within a frozen text-to-image diffusion model to assist 3D affordance grounding. Specifically, the reference image is fed into the frozen diffusion U-Net to extract the diffusion model’s internal features $A_{v}$, and then we use an aggregation network to obtain the affordance priors $A_{g}$ (Sec.~\ref{Unet}, Sec.~\ref{captio}). With the internal features and the text features, DAG leverages the proposed Affordance Blocks to fuse them and obtain affordance tokens $A_{f}$ (Sec.~\ref{affordance_block}). The point cloud features and the [CLS] token are extracted by a pre-trained point encoder and then fed into a lightweight multi-source decoder, along with the affordance tokens $A_{f}$, to obtain the final affordance mask $A_{m}$ (Sec.~\ref{point}, Sec.~\ref{MSAD}). In the following sections, we describe each of these components.

\begin{figure*}[t]
    \centering
    \includegraphics[width=\linewidth]{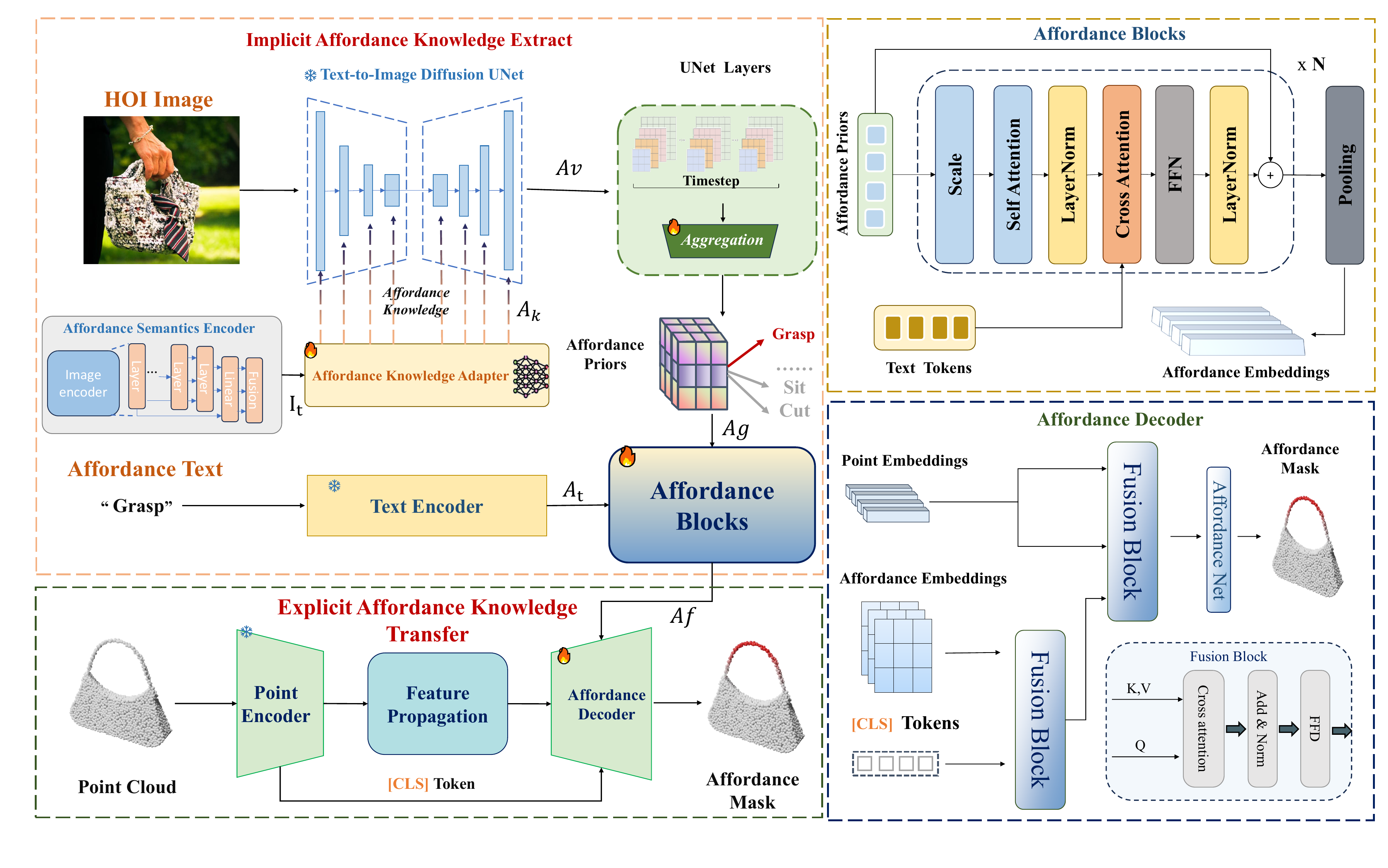}
    \caption{\textbf{DAG pipeline.} Specifically, DAG employs a frozen diffusion U-Net to extract rich affordance knowledge from HOI images. An affordance semantics encoder and an implicit projector are adopted to map the extracted affordance knowledge into the U-Net structure. We then use an aggregation network to construct a unified affordance bank. Afterward, Affordance Blocks fuse the text embeddings with the affordance bank, which are further fed into a multi-modal affordance decoder. Finally, the decoder interacts with 3D point cloud embeddings to generate dense 3D affordance masks.}
    \label{pipeline}
\end{figure*}

\subsection{Affordance Knowledge Extraction and Fusion}
\label{Unet}

Advanced diffusion-based text-to-image generative models~\cite{rombach2022high} use a U-Net architecture to learn the diffusion and denoising processes. The U-Net produces a number of intermediate feature maps that are large and unwieldy, varying in resolution and detail, but contain rich information about texture and semantics. At every step of the denoising process, diffusion models use the text input to infer the denoising direction of the noisy input image. This encourages visual features to correlate with rich, semantically meaningful text descriptions. Thus, the feature maps output by the U-Net blocks can be regarded as rich and dense features for affordance grounding. Our method performs a single forward pass of an input HOI image through the diffusion model to extract its visual affordance representation, as opposed to going through the entire multi-step generative diffusion process. Formally, given an input image $x$, we first sample a noisy image $x_t$ at step $t$ as:
\begin{equation}
    x_t \triangleq \sqrt{\bar{\alpha}_t}\, x + \sqrt{1 - \bar{\alpha}_t}\, \epsilon, \quad \epsilon \sim \mathcal{N}(0, \mathbf{I})
\end{equation}
where $t$ is the diffusion time step we use, $\alpha_1, \ldots, \alpha_T$ represent a pre-defined noise schedule, and $\bar{\alpha}_t = \prod_{k=1}^{t} \alpha_k$, as defined in~\cite{rombach2022high}. Then We utilize the proposed Affordance Semantics Encoder $\mathcal{T}$ (Sec. \ref{captio}) to encode the affordance semantics implicitly and project the semantics into affordance knowledge $A_k$, and finally, we extract the text-to-image diffusion U-Net's internal features by feeding the implicit affordance semantics into the U-Net, which can be formulated as :
\begin{equation}
    A_{v,l} = \mathrm{UNet}(x_t, A_k),
\end{equation}
where $A_{v,l}$ is the U-Net feature of layer $l$. After obtaining the multi-scale features $A_{v}$ from the U-Net, we propose an interpretable aggregation network that learns mixing weights across features to highlight the layers that provide the most useful information for universal affordance knowledge priors $A_{g}$, which can be formulated as:
\begin{align}
    A_g = \sum_{l=1}^{L} w_{l} \cdot A_{v,l},
\end{align}
where $L$ is the number of layers, $w_{l}$ is the weight.

\subsection{Affordance Semantics Encoder}
\label{captio}

To perform the single forward pass, one question is how to obtain the text embeddings for the model. Inspired by previous works~\cite{xu2023open}, we adopt an Affordance Semantics Encoder (ASE) to generate an implicit text embedding from the image itself and input this embedding into the diffusion model directly. Specifically, we use a frozen CLIP image encoder to encode the image and aggregate features from the last layers. Each layer’s features are first processed by a linear projection, and then all features are linearly combined via a weighted summation. Finally, we use a learned MLP to project the implicit affordance knowledge into the U-Net. This can be described as:
\begin{equation}
    I_{t} = \mathcal{T}(x),
\end{equation}
\begin{equation}
    A_{k} = \mathrm{Proj}(I_{t}).
\end{equation}

\subsection{Affordance Blocks}
\label{affordance_block}
To strengthen the generalization ability of our method, we propose an Affordance Block module to fuse the visual affordance knowledge priors $A_{g}$ and the text features $A_{t}$, which are extracted by a frozen CLIP text encoder. As shown in Fig.~\ref{pipeline}, we first perform a scaling operation between $A_{g}$ and $A_{t}$, and then a self-attention mechanism is used to fuse them, followed by residual connections and layer normalization. Next, we inject the implicit affordance text knowledge $A_{t}$ as a condition via a cross-attention operation. Afterward, we use an FFN, residual connections, layer normalization, and an average pooling layer to obtain the affordance embeddings $A_{f}$, which can be formulated as:
\begin{equation}
    A_{r} = \mathrm{AffordanceBlock}(A_{g}, A_{t}),
\end{equation}
\begin{equation}
    A_{f} = \mathrm{Pooling}(A_{r}).
\end{equation}

\subsection{Point Encoder and Propagation}
\label{point}
We utilize a pre-trained 3D encoder to capture point cloud features. To adapt these features for dense 3D prediction, we adopt a geometry-guided upsampling and feature propagation strategy to obtain dense point features from the 3D encoder. More details about the propagation process are provided in the supplementary material.

\subsection{Multi-Source Affordance Decoder}
\label{MSAD}
As shown in Fig.~\ref{pipeline}, we propose a lightweight multi-source affordance decoder, which utilizes the [CLS] token extracted from a frozen pre-trained 3D encoder, affordance embeddings $A_{f}$, and point features ${f}_{\text{p}}$ to obtain the affordance mask $A_{\text{mask}}$. We first fuse the global [CLS] token and affordance embeddings $A_{f}$ via cross-attention:
\begin{equation}
    \mathbf{F}_p =\text{softmax}\left(\frac{\mathbf{Q} \cdot \mathbf{K}^T}{\sqrt{d}}\right) \cdot \mathbf{V} ,
\end{equation}
where $\mathbf{Q}$ represents the global [CLS] token and $\mathbf{K}$ and $\mathbf{V}$ represent the affordance embeddings $A_{f}$. We then apply residual connections and FFN operations and use the resulting features as $\mathbf{Q}$ to perform another fusion process with point features ${f}_{\text{p}}$ as $\mathbf{K}$ and $\mathbf{V}$ to obtain the fused features $\mathbf{P}_f$. Finally, we obtain the affordance mask $A_{\text{mask}}$ by feeding the fused features into an Affordance Net, which is an MLP:
\begin{equation}
    A_{\text{mask}} = \mathrm{MLP}(\mathbf{P}_f).
\end{equation}
\section{Experiment}
\label{Exeperiment}

In this section, we conduct experiments to demonstrate the efficacy of our model. We first introduce the experimental settings (Sec.~\ref{dataset}). Then, we give a detailed description of our implementation (Sec.~\ref{details}) and compare our results against the state-of-the-art (SOTA) models for 3D affordance grounding (Sec.~\ref{Result}). Lastly, we present ablation studies (Sec.~\ref{abla}) to demonstrate the effectiveness of the components and the open-world generalization ability of our method (Sec.~\ref{open-world}).

\subsection{Experimental Settings}
\label{dataset}

\paragraph{Dataset.} To validate the effectiveness and generalization of our method from a comprehensive perspective, we use the PIAD1 and PIAD2 datasets as our benchmarks. Both datasets are well structured and are divided into two parts: seen and unseen.  

\paragraph{Baselines and Metrics.} To provide a comprehensive and effective evaluation, we select several open-sourced cross-modal learning works~\cite{li2021referring,liu2023gres,aiello2022cross,xu2018pointfusion} and open-vocabulary affordance learning works~\cite{nguyen2023open,yang2023grounding,laso,shao2024great,zhu2025grounding}. For evaluation metrics, we follow previous works and finally choose four metrics: Area Under the Curve (AUC)~\cite{lobo2008auc}, Mean Intersection over Union (mIoU)~\cite{rahman2016optimizing}, Similarity (SIM)~\cite{swain1991color}, and Mean Absolute Error (MAE)~\cite{willmott2005advantages}.









\subsection{Implementation Details}
\label{details}
\textbf{Architecture.} We use the stable diffusion model pre-trained on a subset of the LAION~\cite{schuhmann2021laion} dataset as our text-to-image diffusion model. Following ODISE~\cite{xu2023open}, we extract feature maps from every three of its U-Net blocks and resize them to create a feature pyramid. Following ~\cite{xu2023open}, we set the time step used for the diffusion process to $t = 0$. By default, we use the clip text encoder to encode the affordance text. As for point cloud, we use pre-trained Uni3D~\cite{zhang2023uni3d} as our point cloud encoder. During training, we use the Adam~\cite{loshchilov2017decoupled} optimizer with a learning rate set to 1e-4. 

\paragraph{Training objectives.} Our model seeks to transfer the affordance knowledge in the text-to-diffusion model into a 3D affordance decoder. Thus, we solely employ Dice loss~\cite{milletari2016v} and Binary Cross Entropy(BCE) loss~\cite{article} to guide the segmentation mask prediction:

\begin{equation}
    \mathcal{L} = \mathcal{L}_{BCE} + \mathcal{L}_{Dice}.
\end{equation}

Following previous research, we show the comparison results on PIAD~\cite{yang2023grounding}. We split the dataset into two parts, including \textit{Seen} and \textit{Unseen}. \textit{Seen}: This default setting maintains similar distributions of object classes and affordance types across both training and testing phases. \textit{Unseen}: This configuration is specifically designed to evaluate the model's ability to generalize affordance knowledge. In this setup, certain affordance-object pairings are deliberately omitted from the training set but introduced during testing. The training
 is done on one A100 GPU for 80 epochs for the main experiments. And we show the effectiveness of our model by answering the following questions:
\textbf{Q1: How is our model compared to other baselines?} \textbf{Q2: How effective are the proposed components?}
\begin{table}[t]
\caption{\textbf{Main Results.} The overall results of all comparative methods, the best results are in \textbf{bold} and the second results are in \underline{underline}.}

\centering
\resizebox{0.9\linewidth}{!}{

\begin{tabular}{@{}lccccc@{}}

\toprule
& Method & \textit{mIoU}$\uparrow$ & \textit{AUC}$\uparrow$ & \textit{SIM}$\uparrow$ & \textit{MAE}$\downarrow$ \\
\midrule
\multirow{10}{*}{\rotatebox{90}{Seen}} 
& ReferTrans   & 11.32 & 78.89 & 0.497 & 0.129 \\
& ReLA  & 12.08 & 79.13 & 0.483 & 0.125 \\
& PFU  & 12.31 & 77.50 & 0.432  & 0.135 \\
& XMF  & 12.94 & 78.24 &0.441 & 0.127 \\
& IAGNet   & 20.51 & 84.85 & 0.545 & 0.098 \\
& LASO  & 21.14 & 86.12 & 0.559 & 0.092 \\
& GREAT  & \underline{22.72}   & \underline{87.94}  & \underline{0.594} & \underline{0.087} \\

& Ours & \textbf{24.84$\pm$0.5} & \textbf{90.16$\pm$0.3} & \textbf{0.637$\pm$0.03} & \textbf{0.078$\pm$0.01} \\
\midrule
\midrule

\multirow{10}{*}{\rotatebox{90}{Unseen}} 
& ReferTrans   & 7.130 & 67.40 & 0.327 & 0.151 \\
& ReLA  & 7.320 & 68.20 & 0.323 & 0.147 \\
& PFU  & 5.330 & 61.87 & 0.330 & 0.193 \\
& XMF  & 5.680 & 62.58 &0.342 & 0.188 \\

& IAGNet   & 7.950 & 71.84 & 0.352 & 0.127 \\
& LASO  & 8.110 & 71.98 & 0.366 & 0.126 \\
& GREAT  & \underline{8.820}   & \underline{73.61}  & \underline{0.384} & \underline{0.124} \\

& Ours & \textbf{9.730$\pm$0.4} & \textbf{76.69$\pm$0.3} & \textbf{0.414$\pm$0.05} & \textbf{0.120$\pm$0.02} \\

\midrule

\bottomrule
\end{tabular}}

\label{result1}
\end{table}

\begin{table}[t]
\caption{\textbf{Main Results.} The overall results of all comparative methods on the PIAD2 dataset, the best results are in \textbf{bold} and the second results are in \underline{underline}. DAG performs well even on the unseen affordance setting.}
\centering
\resizebox{0.84\linewidth}{!}{

\begin{tabular}{@{}lccccc@{}}

\toprule
& Method & \textit{mIoU}$\uparrow$ & \textit{AUC}$\uparrow$ & \textit{SIM}$\uparrow$ & \textit{MAE}$\downarrow$ \\
\midrule
\multirow{7}{*}{\rotatebox{90}{Seen}} 
& FRCNN  & 33.55 & 87.05 & 0.600 & 0.082 \\
& XMF  & 33.91 & 87.39 &0.604 & 0.078 \\
& IAGNet   & 34.29 & 89.03 & 0.623 & 0.076 \\
& LASO  & 34.88 & 90.34 & 0.627 & 0.077 \\
& GREAT  & \underline{38.03}   & \underline{91.99}  & \underline{0.676} & \underline{0.067} \\

& Ours & \textbf{47.19$\pm$0.3} & \textbf{94.71$\pm$0.4} & \textbf{0.779$\pm$0.02} & \textbf{0.062$\pm$0.02} \\
\midrule

\multirow{7}{*}{\rotatebox{90}{Unseen Obj}} 
& FRCNN  & 18.08 & 72.20 & 0.362 & 0.152 \\
& XMF  & 17.40 & 74.61 &0.361 & 0.126 \\
& IAGNet   & 16.78 & 73.03 & 0.351 & 0.123 \\
& LASO  & 16.05 & 73.32 & 0.354 & 0.123 \\
& GREAT  & \underline{20.16}   & \underline{79.57}  & \underline{0.402} & \underline{0.109} \\

& Ours & \textbf{28.93$\pm$0.7} & \textbf{85.41$\pm$0.3} & \textbf{0.592$\pm$0.05} & \textbf{0.102$\pm$0.02} \\

\midrule

\multirow{7}{*}{\rotatebox{90}{Unseen Aff}} 
& FRCNN  & 7.88 & 58.09 & 0.208 & 0.160 \\
& XMF  & 7.96 & 59.08 &0.210 & 0.156 \\
& IAGNet   & 8.11 & 60.99 & 0.225 & 0.152 \\
& LASO  & 8.37 & 64.07 & 0.228 & 0.140 \\
& GREAT  & \underline{12.05}   & \underline{69.81}  & \underline{0.290} & \underline{0.127} \\

& Ours & \textbf{16.09$\pm$0.4} & \textbf{75.23$\pm$0.3} & \textbf{0.372$\pm$0.05} & \textbf{0.123$\pm$0.02} \\

\midrule
\bottomrule
\end{tabular}}

\label{result2}
\end{table}

\subsection{Main Results(Q1)}
\label{Result}
\begin{table}[t]
\caption{\textbf{Ablation study on the Text Caption.} \textit{Verb} represents the affordance verb text. The best results are in \textbf{bold} and the second results are in \underline{underline}}
\centering
\resizebox{0.9\linewidth}{!}{

\begin{tabular}{@{}lcccccc@{}}
\toprule
& Variants & \textit{mIoU}$\uparrow$ & \textit{AUC}$\uparrow$ & \textit{SIM}$\uparrow$ & \textit{MAE}$\downarrow$ \\
\midrule
& Empty &13.9  &80.5  & 0.462 &0.116  \\
& BLIP &18.4  &85.5  & 0.542 &0.108  \\
& Verb &20.2  &86.7  &0.578 &0.094  \\
& Ours(ASE) & \textbf{24.84$\pm$0.5} & \textbf{90.16$\pm$0.3} & \textbf{0.637$\pm$0.03} & \textbf{0.078$\pm$0.01} \\
\bottomrule
\end{tabular}}

\label{cap}
\end{table}

\begin{figure*}[t]
    \centering
    \includegraphics[width=\linewidth]{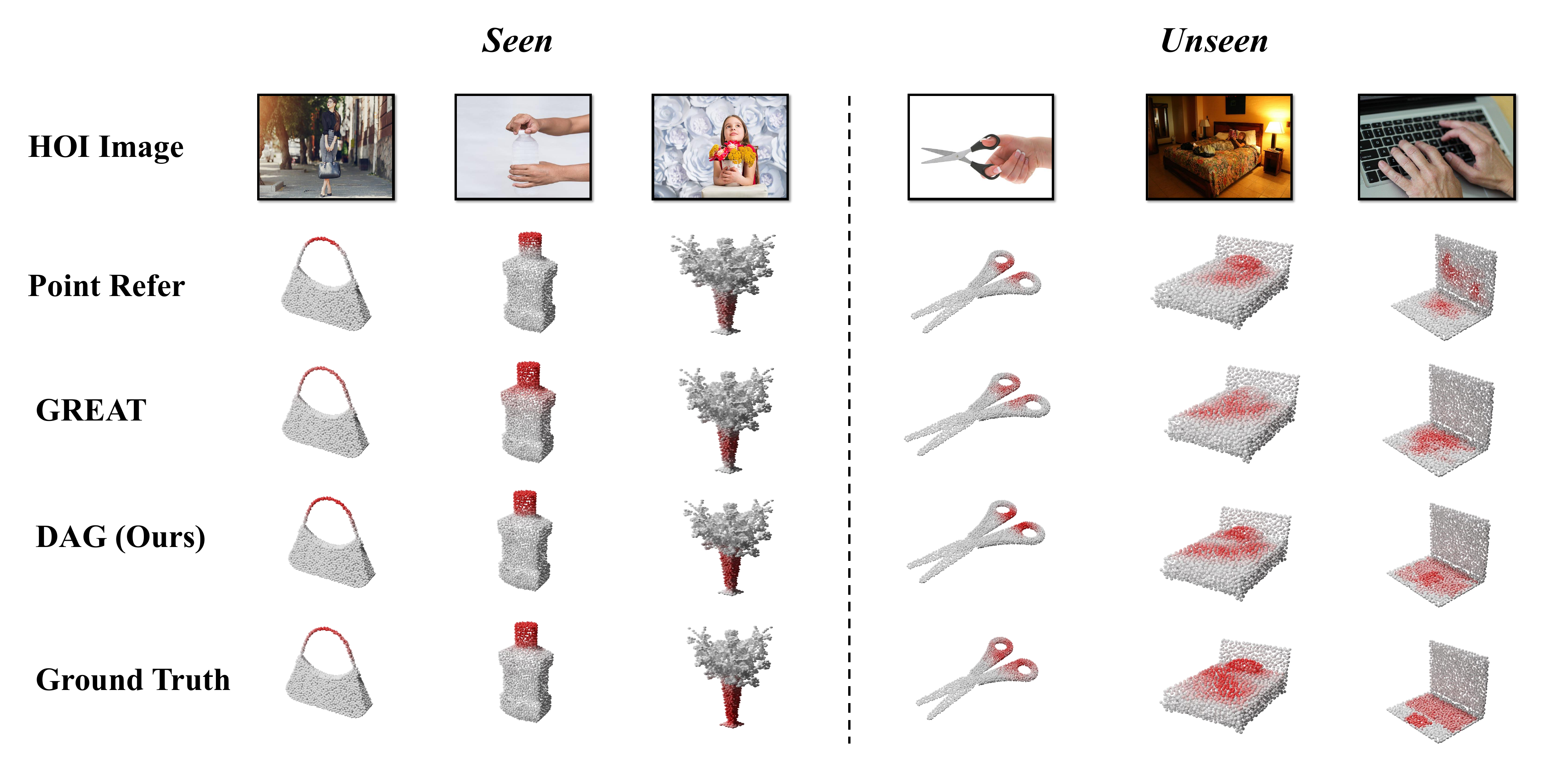}
    \caption{\textbf{Affordance Visualization.} DAG achieves more accurate results in both seen and unseen settings. For more visualization results, please check our Appendix.}
    \label{vis}
\end{figure*}

In Table \ref{result1}, our model demonstrates superior performance across all evaluation metrics compared to the baseline:

\paragraph{DAG vs.Other methods.} To evaluate the generalization capability of our method, we analyze its performance across seen and unseen settings. Our proposed method (DAG) demonstrates superior performance compared to previous methods on both seen and unseen settings. This consistent superiority across datasets validates the robustness and effectiveness of DAG in the open world.

As can be seen in Fig. \ref{vis}, we present a qualitative comparison of affordance predictions under both seen and unseen settings. Our method, DAG, shows strong generalization, extracting the general affordance knowledge within human-object images. And DAG can also maintain consistent and robust performance across diverse scenarios.

\begin{table}[t]
\caption{We investigate the improvement of Affordance Block and [CLS] Token on the model performance based on the baseline. The best results are in \textbf{bold} and the second results are in \underline{underline}}
\resizebox{\linewidth}{!}{

\begin{tabular}{c|cc|cccc}
\toprule
\multicolumn{1}{l|}{} & \textbf{Affordance Block} & \textbf{[CLS]}  & \textit{mIOU}$\uparrow$ &\textit{AUC}$\uparrow$ & \textit{SIM}$\uparrow$ & \textit{MAE}$\downarrow$ \\ 
\midrule
\multirow{4}{*}{\rotatebox{90}{\textbf{\texttt{Seen}}}} & $\times$ & $\times$& 10.02 &  79.95  & 0.398 & 0.124 \\
& \checkmark & $\times$ & 18.56 &  87.04 & 0.582 & 0.111 \\
 &$\times$ & \checkmark & 12.97& 84.52  & 0.470 & 0.116 \\
 & \checkmark & \checkmark  & \textbf{24.84$\pm$0.5} & \textbf{90.16$\pm$0.3} & \textbf{0.637$\pm$0.03} & \textbf{0.078$\pm$0.01} \\
 \midrule
\multirow{4}{*}{\rotatebox{90}{\textbf{\texttt{Unseen}}}} &$\times$ & $\times$ & 8.45 & 67.84 & 0.323 & 0.138 \\
& \checkmark & $\times$ & 9.02& 74.30  & 0.374 & 0.126 \\
 &$\times$ & \checkmark & 8.63& 70.51  & 0.360 & 0.131 \\
 & \checkmark & \checkmark & \textbf{9.730$\pm$0.4} & \textbf{76.69$\pm$0.7} & \textbf{0.414$\pm$0.05} & \textbf{0.120$\pm$0.02} \\
 \bottomrule
\end{tabular}}

\label{ablation1}
\end{table}

\subsection{ Ablation Study and Analysis(Q2)}
\label{abla}
In this section, we conduct a comprehensive ablation study and analysis to investigate the effect of different framework designs.

\paragraph{Effectiveness of Affordance Blocks.} Table. \ref{ablation1} reports the impact on evaluation metrics of the Affordance Blocks. Introducing this module results in a substantial improvement over the baseline, which underscores that our method enables deep integration of high-dimensional semantic features representing instructions with dense features from point clouds and rich affordance world knowledge within the diffusion model, thereby making affordance reasoning more effective.

\paragraph{Effectiveness of [CLS] Token.} The influence of \textbf{[CLS] Token} on evaluation metrics is also shown in Table. \ref{ablation1}. With the integration of the \textbf{[CLS]}-guided mask, results of both settings can be improved, suggesting that the \textbf{[CLS]} token can introduce the global point features, which are essential for affordance grounding.

\paragraph{Different Captioner Strategy.} As shown in the Table. \ref{cap}, we construct several baselines to show the effectiveness of our implicit captioning module: Affordance Semantics Encoder. Since our implicit self-prompt captioning module derives its caption from a text-image discriminative model trained on Internet-scale data, it is able to generalize best among all variants compared.

\paragraph{Different Affordance Extractor.} We conducted comparative experimental analysis on the superiority of using text-to-image diffusion as the affordance extractor. We selected commonly used image feature networks (ResNet), text-image pair pre-trained model (CLIP), and visual self-supervised model (DINOv2) to compare with diffusion. As shown in Table \ref{extractor}, the diffusion affordance extractor outperforms all the variants. As the saying goes, \textit{\textbf{What I can not create, I do not understand}}, the diffusion model can understand both the global semantic concept and the details of the image, which comes from its generation training process. While other models lack semantic concepts (such as ResNet, DINO) or only understand the global information of the image, but lose details (CLIP).

\begin{table}[t]
\caption{\textbf{Ablation study on Affordance Extractor.} We select several representative models to conduct experiments to compare the affordance knowledge extraction ability. DAG (Diffusion-based) outperforms on all metrics.}
\centering
\resizebox{0.84\linewidth}{!}{

\begin{tabular}{@{}lcccccc@{}}
\toprule
& Variants & \textit{mIoU}$\uparrow$ & \textit{AUC}$\uparrow$ & \textit{SIM}$\uparrow$ & \textit{MAE}$\downarrow$ \\
\midrule
& ResNet &13.7  &82.3  &0.514 &0.128  \\
& CLIP &21.2  &86.2  & 0.573 &0.091  \\
& DINOv2 &22.3  &87.5  & 0.582 &0.084  \\
& Ours &\textbf{24.84$\pm$0.5} & \textbf{90.16$\pm$0.3} & \textbf{0.637$\pm$0.03} & \textbf{0.078$\pm$0.01} \\
\bottomrule
\end{tabular}}

\label{extractor}
\end{table}

\begin{wraptable}{r}{0.42\textwidth}
  \centering
\caption{Timestep ablation.}

\begin{tabular}{c|cccc}
\toprule
$t$ & AUC$\uparrow$ & mIoU$\uparrow$ & SIM$\uparrow$ & MAE$\downarrow$ \\
\midrule
0   & \textbf{90.16} & \textbf{24.85} & \textbf{0.637} & \textbf{0.078}\\
50  & 88.92 & 23.11 & 0.619 & 0.082 \\
100 & 88.11 & 21.93 & 0.608 & 0.084 \\
\bottomrule
\end{tabular}
\label{tab:timestep_ablation}
\end{wraptable}

\paragraph{Timestep sensitivity.}
In our default setting, we empirically set the diffusion timestep $t=0$ to acquire high-quality feature representations. To fully validate the rationality and superiority of this key hyperparameter choice, we conduct a comprehensive ablation study on the PIAD1 seen benchmark. For a fair comparison, all experimental configurations are strictly consistent across different timesteps. As quantitatively illustrated in Table~\ref{tab:timestep_ablation}, the model achieves the optimal performance on all evaluation metrics when the timestep is set to 0. In contrast, larger diffusion timesteps will introduce excessive noise interference in the feature extraction process, which blurs detailed texture information, weakens the dense spatial localization capability of the model, and ultimately reduces the accuracy and robustness of prediction results. The above ablation results firmly demonstrate that setting $t=0$ is the most suitable configuration for our diffusion-based feature extraction module.

\subsection{More Analysis about the Open-World Generalization Ability}
\label{open-world}
\paragraph{Performance on Open-World settings.} To further validate the generalization ability of DAG in open-world scenarios, we conduct extensive experiments under the few-shot setting, where the model is trained with very little data, and is expected to recognize unseen objects and affordances during inference. Specifically, we evaluate the model when only a small number of labeled examples are available for each affordance category. As demonstrated in the Table \ref{tab:few_shot}, even under the challenging 1-shot setting, our model still maintains competitive affordance grounding performance, which clearly demonstrates its powerful ability to generalize to novel categories with limited supervision. As the number of shots increases from 1 to 5, the performance of DAG gradually rises and steadily outperforms competing approaches, which are trained with the full dataset, further verifying the effectiveness and robustness of unlocking affordance knowledge from diffusion models for 3D affordance grounding. And we also provid with some few shots experiments results of other methods in the supplementary material.

\begin{figure}[ht]
    \centering
    \includegraphics[width=\linewidth]{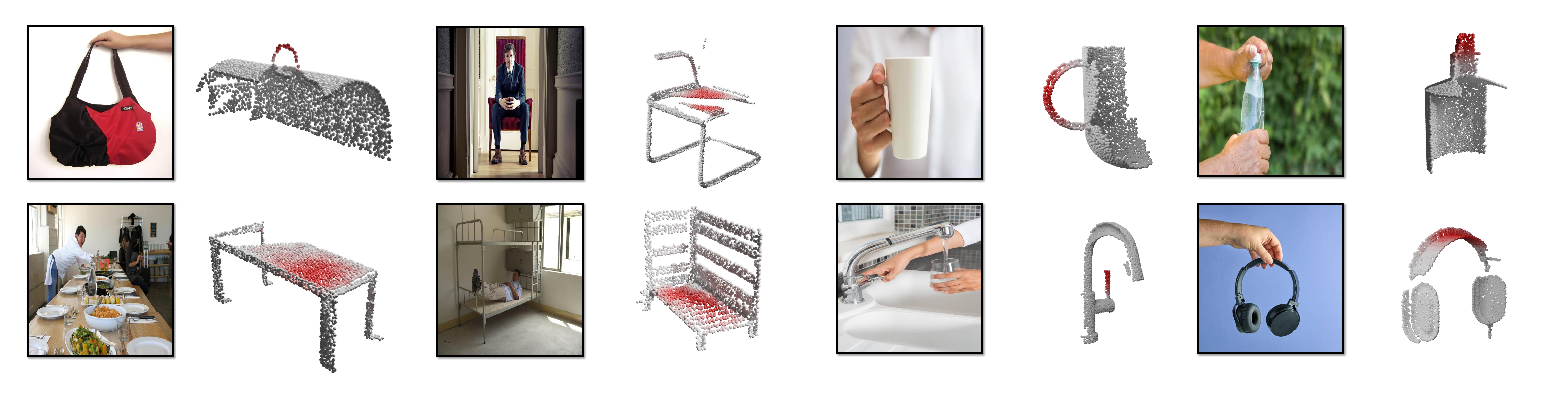}
    \caption{\textbf{Visualization Results on Partial Point Clouds.} Even if the input point cloud is incomplete, DAG can still predict the affordance area well, which shows the strong generalization ability of our method.}
    \label{part_vis}
\end{figure}

\paragraph{Visualization Results on Partial Point Clouds.} In the real physical world, due to the limited viewing angles observed by embodied agents and the limitations of sensor perception, the point clouds they obtain are partially incomplete and with noise. This poses a huge challenge for the practical application of algorithms. We selected incomplete point clouds from the 3D AfferdanceNet dataset to test the algorithm. As shown in Figure \ref{part_vis}, even if the point clouds are partial, DAG can roughly predict the affordance region, demonstrating the powerful generalization applicability.

\begin{table}[t]
\caption{Performance of our DAG under different few-shot settings. DAG can achieve robust and competitive performance on limited and Out-of-Distribution Data, which shows that our model can truly understand the concept of affordance.}
\centering
\resizebox{0.8\linewidth}{!}{

\begin{tabular}{lcccc}
\toprule
Setting & \textit{mIoU}$\uparrow$ & \textit{AUC}$\uparrow$ & \textit{SIM}$\uparrow$ & \textit{MAE}$\downarrow$\\
\midrule
Ours (5 shots) (\color{red}2\% training data) & 15.74 & 82.04 & 0.4869 & 0.121\\

Ours (3 shots) (\color{red}1\% training data) & 14.34  & 80.22 & 0.4955 & 0.124\\
Ours (1 shot) (\color{red}less than 1\% ) & 12.98 & 79.42 & 0.4952 & 0.125\\
PFU (All data) & 12.31 & 77.50 & 0.432  & 0.135 \\

IAGNet (All data)  & 20.51 & 84.85 & 0.545 & 0.098 \\
Ours (All data)  &\textbf{24.84$\pm$0.5} & \textbf{90.16$\pm$0.3} & \textbf{0.637$\pm$0.03} & \textbf{0.078$\pm$0.01} \\
\bottomrule
\end{tabular}}

\label{tab:few_shot}
\end{table}

\section{Conclusion and Future Work}
\label{conclusin}
We present a novel framework, \textbf{DAG}, which is designed to unlock the rich affordance knowledge within frozen text-to-image diffusion models for 3D affordance grounding. By leveraging the frozen U-Net to extract affordance priors in a feature pyramid manner and integrating Affordance Blocks, an Affordance Semantics Encoder, and a multi-source Affordance Decoder, DAG achieves precise affordance prediction, and extensive experiments demonstrate that DAG outperforms previous state-of-the-art methods and show strong open-world generalization ability. Future work will focus on transferring rich affordance knowledge to 3D grounding despite limited 3D data.
\section{Acknowledgement}
This paper is supported by the National Natural Science Foundation of China (No. 62406161).

\bibliographystyle{splncs04}
\bibliography{main}

\end{document}